\documentclass[10pt,twocolumn,letterpaper]{article}

\usepackage{cvpr}
\usepackage{times}
\usepackage{epsfig}
\usepackage{graphicx}
\usepackage{amsmath}
\usepackage{amssymb}
\usepackage{subcaption} 

\usepackage[pagebackref=true,breaklinks=true,letterpaper=true,colorlinks,bookmarks=false]{hyperref}
\pdfoutput=1

\cvprfinalcopy 

\def\cvprPaperID{****} 

\begin{document}

\title{LP-3DCNN: Unveiling Local Phase in 3D Convolutional Neural Networks}

\author{Sudhakar Kumawat and Shanmuganathan Raman\\
Indian Institute of Technology Gandhinagar\\
Gandhinagar, Gujarat, India\\
{\{\tt\small sudhakar.kumawat, shanmuga\}@iitgn.ac.in}
}

\maketitle

\begin{abstract}
Traditional 3D Convolutional Neural Networks (CNNs) are computationally expensive, memory intensive, prone to overfit, and most importantly, there is a need to improve their feature learning capabilities. To address these issues, we propose Rectified Local Phase Volume (ReLPV) block, an efficient alternative to the standard 3D convolutional layer. The ReLPV block extracts the phase in a 3D local neighborhood (e.g., $3\times3\times3$) of each position of the input map to obtain the feature maps. The phase is extracted by computing 3D Short Term Fourier Transform (STFT) at multiple fixed low frequency points in the 3D local neighborhood of each position. These feature maps at different frequency points are then linearly combined after passing them through an activation function.  The ReLPV block provides significant parameter savings of at least, $3^3$ to $13^3$ times compared to the standard 3D convolutional layer with the filter sizes $3\times 3\times 3$ to $13\times 13\times 13$, respectively. We show that the  feature learning capabilities of the ReLPV block are significantly better than the standard 3D convolutional layer. Furthermore, it produces consistently better results across different 3D data representations. We achieve state-of-the-art accuracy on the volumetric ModelNet10 and ModelNet40 datasets while utilizing only 11\% parameters of the current state-of-the-art. We also improve the state-of-the-art on the  UCF-101 split-1 action recognition dataset by 5.68\% (when trained from scratch) while using only 15\% of the parameters of the state-of-the-art. The project webpage is available at \textcolor{red}{https://sites.google.com/view/lp-3dcnn/home}.
\end{abstract}

\section{Introduction}

Over the past few years, research in the area of 2D CNNs has led to unprecedented advances in a number of computer vision tasks such as image classification, semantic segmentation, and image super-resolution. Apart from performance results, 2D CNNs have also made good progress in other complementary areas such as network compression, binarization, quantization, regularization, etc. Unfortunately, unlike their 2D counterparts, 3D CNNs have not enjoyed the same level of performance jumps on the problems in their domain e.g., video classification and progress in the above mentioned complementary areas. Recent works such as \cite{xie2018rethinking} and \cite{hara2018can}, list down some of the fundamental  barriers in modeling and training of deep 3D CNNs such as (1) they are computationally very expensive, (2) they result in large model size, both in terms of memory usage and disk space, (3) they are prone to overfitting, due to a large number of parameters, (4) and there is a need to improve their feature learning capabilities which may require fundamental changes to their network architecture or the standard 3D convolutional layer \cite{xie2018rethinking,citation-0,tran2018closer}. Despite the above challenges, the current trend in the literature of deep 3D CNNs  is to train computationally expensive, memory intensive, and  very deep networks in order to achieve state-of-the-art results \cite{brock2016generative,diba2018spatio,hara2018can}. 

In this work,  we take a detour from this trend by proposing an alternative to the fundamental building block of the 3D CNNs, the 3D convolutional layer, which is the primary source of high space-time complexity in 3D CNNs. More precisely, we propose Rectified Local Phase Volume (ReLPV) block, an efficient alternative to the standard 3D convolutional layer in 3D CNNs. The ReLPV block comprises of a local phase module, the ReLU activation function and a set of trainable linear weights. The local phase module extracts the local phase information by computing 3D Short Term Fourier Transform (STFT) \cite{hinman1984short} (at multiple low frequency points) in a local $n\times n\times n$ (e.g., $3\times 3\times 3$) neighborhood/volume of each position of the input feature map. The output of the local phase module is then passed through the ReLU activation function in order to obtain the activated response maps of the local phase information at the fixed low frequency points. Finally, a set of trainable linear weights computes the weighted combinations of these activated response maps. The ReLPV block provides significant parameter savings along with computational and memory savings. The ReLPV block based 3D CNNs have much lower model complexity and are less prone to overfitting. Most importantly, its feature learning capabilities are significantly better than the standard 3D convolutional layer. 

Our major contributions in this work are as follows.
\begin{itemize}
	\item We propose ReLPV block, an efficient alternative to the standard 3D convolutional layer. The ReLPV block significantly reduces the number of trainable parameters, at least $3^3$ to $13^3$ times compared to the standard 3D convolutional layer with the filter sizes $3\times 3\times 3$ to $13\times 13\times 13$, respectively.
	\item We show that the ReLPV block achieves consistently better results on different 3D data representations.  We show this on the volumetric ModelNet10 and ModelNet40 datasets by achieving state-of-the-art accuracy using just 11\% parameters of the current state-of-the-art. Moreover, we provide results on the spatiotemporal image sequences. In particular, on the UCF-101 split-1 action recognition dataset, improving the current state-of-the-art by 5.68\% while using just 15\% parameters of the state-of-the-art.    
	\item We present detailed ablation and performance studies of the proposed ReLPV block by varying its various hyperparameters. The analysis will be beneficial for designing ReLPV block based 3D CNNs in future.
\end{itemize}


\section{Related Work}\label{sec:related_work}

Recently, 2D CNNs have achieved state-of-the-art results in most of the computer vision problems \cite{goodfellow2016deep}. Moreover, they have also made significant progress in other complementary areas such as  network compression \cite{howard2017mobilenets, zhang2017shufflenet}, binarization \cite{courbariaux2015binaryconnect,courbariaux2017binarynet,rastegari2016xnor,juefei2017local}, quantization \cite{zhou2017incremental,hubara2017quantized}, regularization \cite{cogswell2015reducing,DBLP:journals/corr/abs-1804-08450,xiong2016regularizing,rodriguez2016regularizing}, etc. Therefore, not surprisingly, there have been many recent attempts to extend this success to the problems in the domain of 3D CNNs e.g., video classification \cite{asadi2017survey}, 3D object recognition \cite{maturana2015voxnet,brock2016generative} and MRI volume segmentation \cite{milletari2016v,cciccek20163d}. Unfortunately, 3D CNNs are computationally expensive and require large memory and disk space. Furthermore, they overfit very easily owing to the large number of parameters involved. Therefore, there has been recent interest in more efficient variants of 3D CNNs.

Inspired from the progress of network binarization techniques in 2D CNNs such as BinaryConnect \cite{courbariaux2015binaryconnect}, BinaryNet \cite{courbariaux2017binarynet}, and XNORNet \cite{rastegari2016xnor}, Ma \emph{et al.} in \cite{citation-0} introduced BV-CNNs, where they fully binarized some of the state-of-the-art 3D CNN models introduced for recognizing voxelized 3D CAD models from the ModelNet datasets \cite{wu20153d}. The binarized version of the 3D CNNs saves significant computation and memory requirements when compared to the floating point baselines. However, this comes at the cost of reduced performance. Furthermore, the binarized network takes binarized inputs only which restricts its application for other 3D data representations such as video classification.    

Another way to reduce the model complexity of 3D CNNs is to replace the 3D convolutions with separable convolutions. This technique has been explored recently in a number of 3D CNN architectures proposed for the task of video classification. The idea of separable convolutions is to first convolve spatially in 2D and then convolve temporally in 1D. This factorization is similar in spirit to the depth-wise separable convolutions used in \cite{xie2017aggregated},  except that here the idea is to apply it to the temporal dimension instead of the feature dimension. The idea has been used in a variety of recent works, including R(2+1)D networks\cite{tran2018closer}, separable-3D CNNs \cite{xie2018rethinking}, Pseudo-3D networks  \cite{qiu2017learning}, and  factorized spatio-temporal CNNs \cite{sun2015human}. The 3D CNNs based on the idea of separable convolutions achieve competitive results compared to the state-of-the-art on the task of video classification at a reduced space-time complexity.

\begin{figure*}[t]
	\begin{center}
		\includegraphics[]{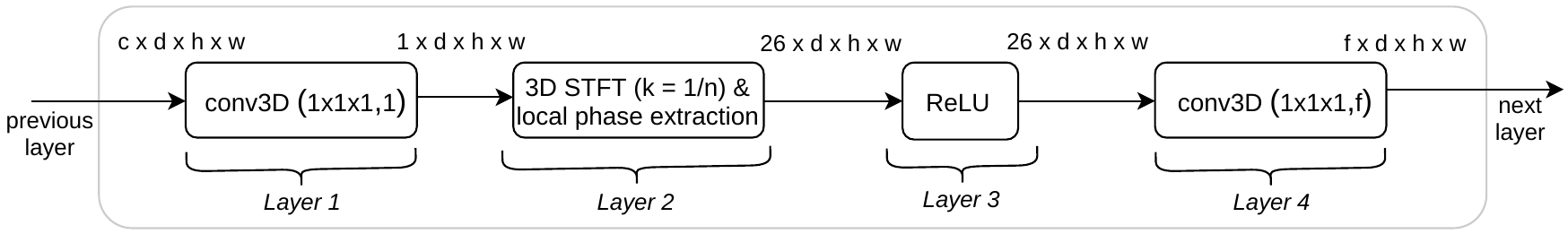} 
		\caption{\textbf{The ReLPV block architecture.}}
		\label{fig:relpv}
	\end{center}
\end{figure*}

\section{Method}\label{sec:method}
\noindent\textbf{Notation.} We denote the feature map output by a layer in a 3D CNN network with the tensor $\mathcal{I}\in \mathbb{R}^{c\times d \times h \times w}$  where $h$, $w$, $d$, and $c$ are the height, width, depth, and number of channels of the feature map, respectively.

\noindent\textbf{The ReLPV Block Architecture.} The ReLPV block is a four-layer alternative representation of the standard 3D convolutional layer. Fig.~\ref{fig:relpv} illustrates the architecture of the ReLPV block.

\noindent\textit{Layer 1.} This layer is the standard 3D convolutional layer with a single filter of size $1\times1\times1$. It takes a feature map of size $c\times d\times h\times w$ as input from the previous layer, and converts it into a single channel feature map of size $1\times d\times h\times w$.  This layer prepares the input for the 3D STFT operation which is computed in  \textit{Layer 2}. Let $f(\textbf{x})$ be  the feature map output of \textit{Layer 1} with size $1\times d\times h\times w$. Here, \textbf{x} is a variable  denoting positions on the feature map  $f(\textbf{x})$. 



\noindent\textit{Layer 2.}
Local phase has been successfully used in images to detect edges and contours for feature extraction \cite{kovesi1999image}. Phase represents the local coherence of different spatial frequencies. Edges and skeletons in image are expressed by their coherence and play a significant role in image understanding \cite{zachevsky2018modelling}. Same property holds true for 3D data representations too. e.g., videos \cite{paivarinta2011volume}. There are many methods for extracting local phase in multiple dimensions \cite{heikkila2009methods}. Our method is inspired from \cite{paivarinta2011volume}. \textit{Layer 2} extracts the local phase spectra of $f(\textbf{x})$ by computing the 3D Short Term Fourier Transform (STFT) in a local $n\times n\times n$ neighborhood $\mathcal{N}_\textbf{x}$ at each position $\textbf{x}$ of  $f(\textbf{x})$ using Equation~\ref{eq:stft}.  
\begin{equation}\label{eq:stft}
F(\textbf{v},\textbf{x})= \sum_{\textbf{y}\in \mathcal{N}_\textbf{x}} f(\textbf{x}-\textbf{y})\exp^{-j2\pi \textbf{v}^T \textbf{y}}
\end{equation}
\noindent Here, $\textbf{v}\in\mathbb{R}^3$ is a frequency variable and $j=\sqrt{-1}$. Using vector notation \cite{jain1989fundamentals}, we
can rewrite Equation~\ref{eq:stft} as shown in Equation~\ref{eq:stftvec}.
\begin{equation}\label{eq:stftvec}
F(\textbf{v},\textbf{x})= \textbf{w}^T_\textbf{v}\textbf{f}_\textbf{x}
\end{equation}
\noindent Here, $\textbf{w}_\textbf{v}$ is the basis vector of the 3D STFT at frequency variable $\textbf{v}$ and $\textbf{f}_\textbf{x}$ is a vector containing all the positions from the neighborhood $\mathcal{N}_\textbf{x}$. Note that, due to the separability of the basis functions, 3D STFT can be computed efficiently for all the positions $\textbf{x}$ in $f(\textbf{x})$ by using simple 1D convolutions for each dimension.  In this work, we consider 13 lowest non-zero frequency variables which are defined as below. 
\begin{table}[htbp]
	\small
	\begin{tabular}{lll}
		$ \textbf{v}_1=[k,0,0]^T $, & $\textbf{v}_2=[k,0,k]^T$, & $\textbf{v}_3=[k,0,-k]^T$,\\ 
		$ \textbf{v}_4=[0,k,0]^T$ , & $\textbf{v}_5=[0,k,k]^T$, & $\textbf{v}_6=[0,k,-k]^T$,\\
		$ \textbf{v}_7=[k,k,0]^T$, & $\textbf{v}_8=[k,k,k]^T$, & $\textbf{v}_9=[k,k,-k]^T$,\\
		$ \textbf{v}_{10}=[k,-k,0]^T$,& $\textbf{v}_{11}=[k,-k,k]^T$,& $\textbf{v}_{12}=[k,-k,-k]^T$, \\ 
		$\textbf{v}_{13}=[0,0,k]^T$,& \multicolumn{2}{l}{where $k = 1/n$}
	\end{tabular}
\end{table}
\begin{figure}[htbp]
	\begin{center}
		\includegraphics[width=0.8\columnwidth, height=0.6\columnwidth]{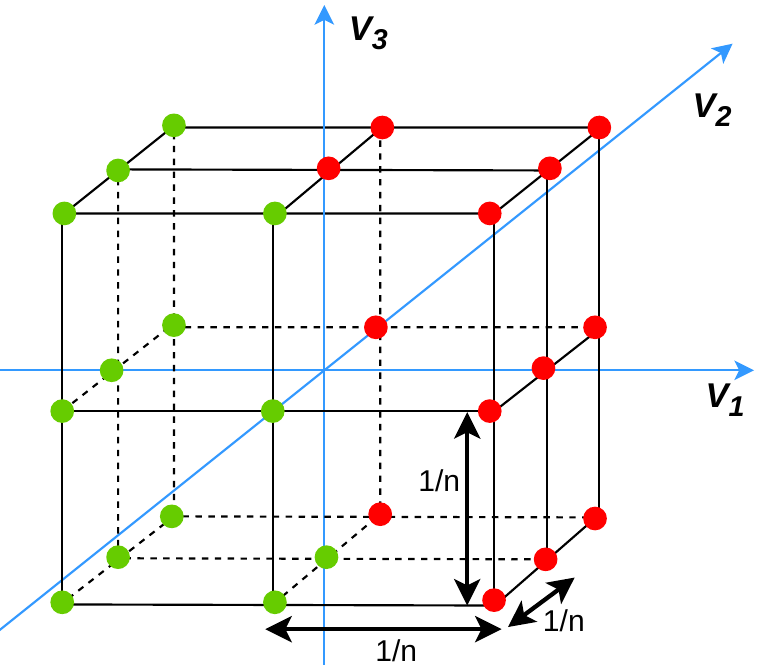} 
		\caption{\textbf{Frequency points used to compute the 3D STFT.} The selected frequency points are marked as red dots. The other frequency points in the green dots are ignored, as they are the complex conjugates of the selected ones.}
		\label{fig:3d-stft}
	\end{center}
\end{figure}
The selected frequency variables are shown as red dots in Fig.~\ref{fig:3d-stft}. Low frequency variables are used because they usually contain most of the information, and therefore they have
better signal-to-noise ratio than the high frequency components \cite{heikkila2009methods}. Let
\begin{equation}\label{eq:stft_3}
\textbf{W} = [\Re\{\textbf{w}_{\textbf{v}_1}, \textbf{w}_{\textbf{v}_2}, \ldots, \textbf{w}_{{\textbf{v}_{13}}}\}, \Im\{\textbf{w}_{\textbf{v}_1}, \textbf{w}_{\textbf{v}_2}, \ldots, \textbf{w}_{{\textbf{v}_{13}}}\}]^T
\end{equation}
\noindent Here, $\textbf{W}$ is a $26\times n^{3}$ transformation matrix corresponding to the 13 frequency variables. $\Re{\{\cdot\}}$ and $\Im{\{\cdot\}}$ return the real and the imaginary parts of a complex number, respectively.  Hence, from Equation~\ref{eq:stftvec} \&~\ref{eq:stft_3}, the vector form of 3D STFT for all the 13 frequency points $\textbf{v}_1$, $\textbf{v}_2$, \ldots ,$\textbf{v}_{13}$ can be written as shown in Equation~\ref{eq:2d_stft}.
\begin{equation}\label{eq:2d_stft}
\textbf{F}_\textbf{x} = \textbf{Wf}_\textbf{x}
\end{equation}
%
\noindent Since, $\textbf{F}_\textbf{x}$ is computed for all positions $\textbf{x}$ of the input $f(\textbf{x})$, it results in an output feature map with size $26\times d\times  h\times w$. A more detailed mathematical formulation of \textit{Layer 2} is provided in Section~\ref{sect:supp}.

\noindent\textit{Layer 3.} Applying non-linearity to the local phase information enables the network to learn complex representations. This layer creates activated response maps of the feature maps obtained from \textit{Layer 2} by using an activation function. We use the ReLU activation function for better efficiency and faster convergence \cite{nair2010rectified}.

\noindent\textit{Layer 4.} This layer is the standard 3D convolutional layer with $f$ filters each of size $1\times 1\times 1$ which takes a feature map of size $26\times d\times h\times w$ as input from \textit{Layer 3} and outputs a feature map of size $f\times d\times h\times w$. Note that, \textit{Layer 1} and \textit{4} get learned during the training phase of the 3D CNN. 

We shall use the notation $\text{ReLPV}(n,f)$ for the ReLPV block, where $n$ and $f$ are its hyperparameters. Here $n$ denotes the size of the local 3D neighborhood from \textit{Layer 2} and $f$ is the number of $1\times1\times1$ filters used in \textit{Layer 4}.

\noindent\textbf{Importance of using STFT and Local Phase.} STFT in multidimensional space was first studied by Hinman \emph{et al.} in \cite{hinman1984short} as an efficient tool for image encoding. It has two important properties which make it useful for our purpose: (1)  Natural images are often composed of objects with sharp edge features. It has been observed that the Fourier phase information accurately represents these edge features. Since STFT in 3D space is simply a windowed Fourier transform, the same property applies \cite{hinman1984short}. Thus, the local phase has the ability to accurately capture the local features in the same way as done by the convolutional filters. (2) STFT decorrelates the input signal \cite{hinman1984short}. Regularization is key for deep learning since it allows training of more complex models
while keeping lower levels of overfitting and achieves better generalization. Decorrelation of features, representations, and hidden activations  has been an active area of research for better regularization of deep neural nets, with a variety of novel regularizers proposed such as DeCov \cite{cogswell2015reducing}, Decorrelated Batch Normalization (DBN) \cite{DBLP:journals/corr/abs-1804-08450}, Structured Decorrelation Constraint (SDC) \cite{xiong2016regularizing} and OrthoReg \cite{rodriguez2016regularizing}. As STFT decorrelates the input representations and due to the reduced number of learnable parameters, the ReLPV block based 3D CNNs are less prone to overfitting and generalize better (for results see Section~\ref{sec:overfit}).

\noindent\textbf{Forward-Backward Propagation in the ReLPV Block.} The end-to-end training of a 3D CNN network  with the ReLPV blocks instead of the standard 3D convolutional layers is straightforward. The steps of forward and backward propagation through the \textit{Layers 1}, \textit{3}  and \textit{4} of the ReLPV block are standard operations in all deep learning libraries. Back propagation in the \textit{Layer 2} is similar to propagating gradients through layers without learnable parameters (e.g. Add, Multiply etc.) as it involves applying the fixed basis matrix \textbf{W} to the input. Note that, during training, only the  $1\times1\times1$ filters in  \textit{Layers 1} and \textit{4} are updated while the weights in the matrix \textbf{W} \emph{} remain unaffected.


\begin{figure*}[t]
	\begin{subfigure}[b]{\columnwidth}
		\centering
		\includegraphics[width=.65\columnwidth, height=0.45\columnwidth]{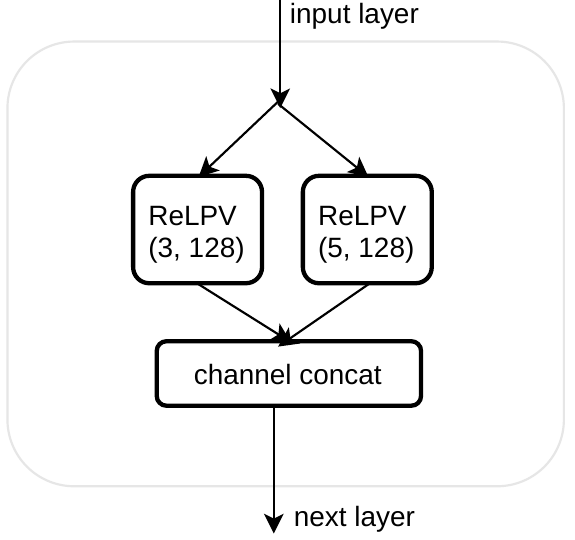}
		\caption{Block 1}
		\label{fig:a}
	\end{subfigure}%
	\begin{subfigure}[b]{\columnwidth}
		\centering
		\includegraphics[width=.65\columnwidth, height=0.45\columnwidth]{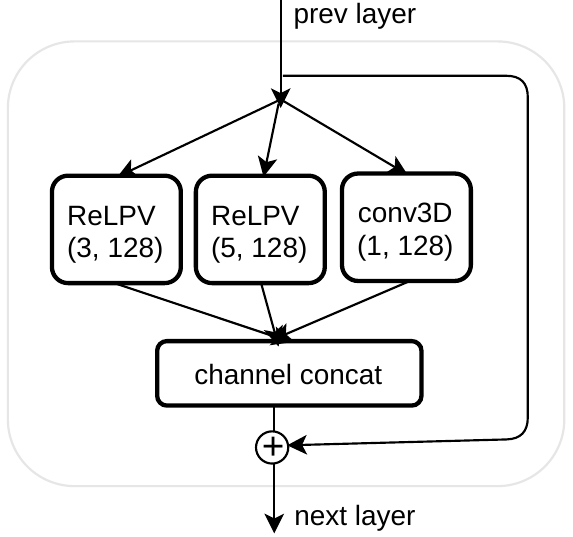}
		\caption{Block 2}
		\label{fig:b}
	\end{subfigure}		
	
	\begin{subfigure}[b]{\textwidth}
		\centering
		\includegraphics[]{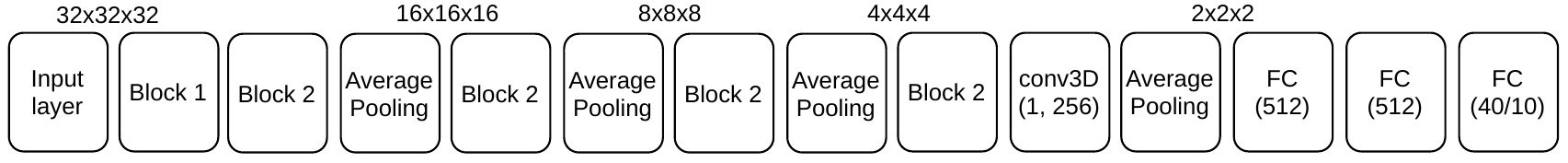}
		\caption{LP-3DCNN network architecture}
		\label{fig:voxception_block}
	\end{subfigure}
	\caption{\textbf{Experiments and comparison with the state-of-the-art}. LP-3DCNN network building blocks and architecture. }\label{fig:blocks}
\end{figure*}



\noindent\textbf{Parameter analysis of the ReLPV Block.} The ReLPV block uses significantly less trainable parameters when compared to the standard 3D convolutional layer with the same filter size/volume and number of input-output channels. Consider a standard 3D convolutional layer with $c$ input  and $f$ output channels. Let $n\times n\times n $ be the size/volume of the filters. Thus, the total number of trainable parameters in a standard 3D convolutional layer is  $c \cdot n^{3} \cdot f$. An ReLPV block with  $c$ input channels and $f$ output channels consists of just $c \cdot 1 + f\cdot 26 $  trainable parameters. Thus, the ratio of the number of trainable parameters in a standard 3D convolutional layer and the proposed ReLPV block is calculated as below.
\begin{equation}
\dfrac{\text{\# params. in 3D conv. layer}}{\text{\# params. in ReLPV block}}= \dfrac{c \cdot n^{3} \cdot f}{c \cdot 1 + f\cdot 26}
\end{equation}
\noindent For simplicity, let us assume $f = c$, i.e., the number of input and output channels are same. Furthermore, in practice, in most deep 3D CNNs $f \geq 27$. Therefore, let $f=27$. This reduces the above ratio to $n^3$. Thus, for a filter of size $3\times 3\times 3$ in the standard 3D convolutional layer, the ReLPV block uses 27 times less trainable parameters. Therefore, numerically, ReLPV block saves atleast $27\times$, $125\times$, $343\times$, $729\times$, $1331\times$,  and $2197\times$ parameters during learning for $3\times3\times 3$, $5\times5 \times 5$, $7\times7\times 7$, $9\times9 \times 9$, $11\times11 \times 11$, and $13\times13 \times 13$ 3D convolutional filters, respectively.

\section{Experiments}\label{sec:experiments}


In this section, we show that the proposed ReLPV block produces consistently better results on different 3D data representations compared to the standard 3D convolutional layer. We demonstrate this on voxelized 3D CAD models and on spatiotemporal image sequences.

\subsection{Experiments and Results on 3D CAD models}
\noindent\textbf{Datasets.} ModelNet \cite{wu20153d} is a large 3D repository of clean CAD models (shapes). The ModelNet10 with 4,899 shapes (train: 3991, test: 908) and ModelNet40 with 12,311 shapes (train: 9843, test: 2468) are commonly used as benchmarking datasets and
consist of 10 and 40 categories, respectively.  Each model is aligned to a canonical frame and then rotated at 12 and 24 evenly-sampled orientations about the z-axis (Az$\times$12 and Az$\times$24 augmentation). These rotated models are then voxelized to a $32\times32\times32$ grid. We use the voxelized versions of \cite{maturana2015voxnet}. The task here is to classify a given voxelized 3D model into its corresponding class.

\subsubsection{ModelNet: Comparison with the baselines}\label{sect:baselines}
\noindent\textbf{Baselines.} We start our experiments by replacing the standard 3D convolutional layer with the proposed ReLPV block (with skip connections) in the baseline networks VoxNet \cite{maturana2015voxnet}, VoxNetPlus \cite{citation-0} and LightNet \cite{zhi2017lightnet}, and call these new networks as LP-VoxNet, LP-VoxNetPlus, and LP-LightNet, respectively. Here LP stands for Local Phase. The standard 3D convolutional layer is replaced with the ReLPV block in a straightforward manner. For example the VoxNet network \cite{maturana2015voxnet} has the following architecture: $\text{conv3D}(5,32,2)-\text{conv3D}(3,32,1)-\text{MP}(2)-\text{FC}(128)-\text{FC}(K)$. Here, $\text{conv3D}(n,f,s)$ is the standard 3D convolutional layer with $f$ filters each of size $n\times n\times n$ applied with stride $s$. MP denotes Max Pooling. FC stands for fully connected layer. $K$ is the number of classes. The equivalent local phase version of VoxNet is: $\text{ReLPV}(5,32,2)-\text{ReLPV}(3,32,1)-\text{MP}(2)-\text{FC}(128)-\text{FC}(K)$. In our earlier discussion on the architecture of the ReLPV block, we focused only on the important hyperparameters and did not discuss other hyperparameters that are commonly used in the standard 3D conv layer, such as the stride information. Such information can easily be incorporated in the ReLPV architecture. Similar procedure is followed while preparing LP-VoxNetPlus and LP-LightNet networks.

\noindent\textbf{Training.} We train these new networks using SGD as optimizer with momentum 0.9 and categorical crossentropy as loss.  During training, we start with a learning rate of 0.008 and reduce it by a factor of 2 if the validation loss plateaus. For LP-VoxNet and LP-VoxNetPlus networks, following \cite{maturana2015voxnet,citation-0}, we first train them on ModelNet40 and then fine-tune on ModelNet10. The opposite is done on LP-LightNet network as done in \cite{zhi2017lightnet}. Following \cite{maturana2015voxnet,citation-0,zhi2017lightnet}, all networks were trained on 12 evenly-sampled rotations of each instance about the z-axis ($\text{Az}\times12$ augmentation). No data augmentation was done on the test data.

\noindent\textbf{Results.} Table~\ref{tab:lpv_baseline} presents the comparison of the new networks with their corresponding baselines. We also compare the new networks with the binarized version of the baselines \cite{citation-0} (as described in section~\ref{sec:related_work}). The local phase version clearly outperforms the corresponding baselines and their binarized versions on both the ModelNet10 and the ModelNet40 datasets.

\begin{table}[t]   
	\centering
	\scriptsize
	\begin{tabular}[width=.9\textwidth]{l|c|c}
		\hline
		\hline
		\textbf{Network} 	& \textbf{ModelNet40 (\%)} 	& \textbf{ModelNet10 (\%)}  \\
		\hline
		
		VoxNet \cite{maturana2015voxnet} (baseline) 		& 83		& 92	\\
		Binary VoxNet \cite{citation-0} & 81.63		& 90.69	\\
		LP-VoxNet (ours) 					& \textbf{86.26}	& \textbf{92.24}			 	\\
		\hline
		VoxNetPlus \cite{citation-0} (baseline) 	& 83.91		& 93.36	 \\
		Binary VoxNetPlus \cite{citation-0} & 85.47		& 92.32	 \\
		LP-VoxNetPlus (ours)  			& \textbf{88.1}		& \textbf{93.4}	 \\
		\hline
		LightNet \cite{zhi2017lightnet} (baseline) 		& 86.90		& \textbf{93.39}	\\
		Binary LightNet \cite{citation-0}     & 84.24  & 92.36  \\
		LP-LightNet (ours)  			& \textbf{87.5}		& 92.95	 \\
		\hline
	\end{tabular}
	\caption{\textbf{Comparison of the baseline networks with the  Local Phase and binarized versions}. The Local Phase version outperforms the baseline and their binarized versions.}
	\label{tab:lpv_baseline}
\end{table}

\subsubsection{ModelNet: Comparison with the state-of-the-art}
\noindent\textbf{Network Architecture.} We follow ideas from the Voxception-ResNet (VRN) architecture  of \cite{brock2016generative} which adopts a simple inception-style architecture with ResNet-style skip connections. The intuition behind this design is to have a maximum number of possible pathways for information to flow through the network. For the first non-downsampling block that follows the input layer (Fig.~\ref{fig:a}), we concatenate an equal number (128) of feature maps from two ReLPV blocks with different local phase volume sizes ($3\times3\times3$ and $5\times5\times5$). For other non-downsampling blocks, we augment the above structure with an additional $1\times1\times1$ convolutional layer that outputs the same number (128) of feature maps as the ReLPV blocks and concatenate it with the other feature maps as shown in Fig.~\ref{fig:b}.  This architecture allows the network to choose between taking a weighted average of the feature maps in the previous layer (i.e. by heavily weighting the $1\times1\times1$ convolutions) or focusing on local phase information (i.e., by heavily weighting the ReLPV blocks). Along with this, skip connections are added as shown in Fig.~\ref{fig:b} for smoother flow of the gradients to the previous layers. For downsampling, we use average pooling with pool size 2 and stride 2. Our final model is shown in Fig.~\ref{fig:voxception_block} with five non-downsampling blocks, followed by two fully connected layers each of size 512, and a final softmax layer for classification. All non-downsampling layers (after batch normalization) and fully connected layers are followed by the ReLU activation function. The layer conv3D(1, 256) is used after the final non-downsampling layer to reduce the number of parameters in the fully connected layers.

\begin{table*}[t]   
	\centering
	\footnotesize
	\begin{tabular}[width=.9\textwidth]{l|c|c|c|c|c}
		\hline
		\hline
		\textbf{Network} & \textbf{Framework}& \textbf{Augmentation} & \textbf{Parameters (Millions)}	& \textbf{ModelNet40 (\%)}	& \textbf{ModelNet10 (\%)}  \\
		\hline
		
		3D ShapeNets \cite{wu20153d} & Single, Volumetric & Az$\times$12 & $\approx$ 38 & 77 & 83.5 \\
		Beam Search \cite{xu2016beam} & Single, Volumetric & - & $\approx$ 0.08 & 81.26 & 88 \\
		3D-GAN \cite{wu2016learning}& Single, Volumetric & - & $\approx$11 &83.3 & 91 \\
		VoxNet \cite{maturana2015voxnet}& Single, Volumetric & Az$\times$12 & $\approx$ 0.92 &83 & 92	\\
		LightNet \cite{zhi2017lightnet}& Single, Volumetric & Az$\times$12 & $\approx$ 0.30 &86.90 & 93.39 \\
		ORION \cite{sedaghatorientation}& Single, Volumetric & Az$\times$12 & $\approx$ 0.91 &- &  93.8 \\
		VRN \cite{brock2016generative}& Single, Volumetric & Az$\times$24 &$\approx$ 18 & 91.33 & 93.61 \\
		\hline
		LP-3DCNN (ours)   & Single, Volumetric & Az$\times$12 & $\approx$ 2 & 89.4 & 93.76   \\
		LP-3DCNN (ours)   & Single, Volumetric & Az$\times$24 & $\approx$ 2 & \textbf{92.1} & \textbf{94.4}   \\
		\hline
		FusionNet \cite{hegde2016fusionnet}& Ensemble, Vol.+ Mul.& (Az, El)$\times$60 & $\approx $118 &90.8 & 93.11 \\
		VRN Ensemble \cite{brock2016generative}& Ensemble, Volumetric& Az$\times$24 & $\approx$ 108 & 95.54 & 97.14 \\
		
		\hline
	\end{tabular}
	\caption{ \textbf{Performance results on the ModelNet datasets}. Az stands for azimuth rotation and El stands for elevation rotation. ``-" means
		that information is not provided for the item in the paper. Vol. stands for volumetric, Mul. stands for multi-view.}
	\label{tab:results}
\end{table*}

\noindent\textbf{Training and Testing.} The input to our network are voxels of size  $32\times32\times32$ from the ModelNet datasets. Following \cite{brock2016generative}, we change the binary voxel range from \{0,1\} to \{-1,5\} to encourage the network to pay more attention to positive entries. Network is trained using SGD as optimizer with momentum 0.9 and categorical crossentropy as loss.  During training, we start with a learning rate of 0.008 and reduce it by a factor of 5 if the validation loss plateaus. All weights are initialized using orthogonal initialization. The network is first trained on the $\text{Az}\times12$ augmented data, then it is fine-tuned on the $\text{Az}\times24$ augmented data at low learning rate.  No data augmentation was done on the test data. Apart from rotations, the data is augmented by adding noise, random translations and horizontal flips to each training example, as done in \cite{maturana2015voxnet,brock2016generative}.

\noindent\textbf{Results.} Table~\ref{tab:results} compares our results with other methods that use voxelized/volumetric ModelNet datasets as input. In order to make a fair comparison, we only consider volumetric network frameworks in this work. We do not include multi-view networks or point cloud-based networks. In single network framework, our proposed network outperforms all the previous networks on both the ModelNet10 and the ModelNet40 datasets. Furthermore, it uses just 2 million parameters compared to the current state-of-the-art, the VRN network, that uses 18 million parameters. In the ensemble framework, the VRN achieves the best performance on both the ModelNet10 and ModelNet40 datasets. However, it has the most complex network architecture with up to 45 layers and 108 million parameters, taking almost 6 days to train. In ensemble framework, our network outperforms FusionNet \cite{hegde2016fusionnet} while using almost 59 times less parameters and significantly less data augmentation.


\subsection{Experiments and Results on Spatiotemporal Image Sequences}
\noindent\textbf{Dataset.}  We use the UCF-101 split-1 action recognition dataset \cite{soomro2012ucf101}. The dataset has been used as a benchmark dataset in \cite{tran2015learning,tran2017convnet,diba2018spatio} for the performance studies and for searching 3D CNN network architectures and hyperparameters for action recognition tasks.

\noindent\textbf{Baseline.} We use the experimental 3D CNN network  proposed by \cite{tran2015learning} for action recognition as baseline which is a smaller version of the C3D network \cite{tran2015learning}. For simplicity, we call this network as mini C3D network or mC3D. The mC3D network with filter size $n\times n\times n$ denoted as $\text{mC3D}_n$  has the following architecture: $\text{conv3D}(n,64)-\text{MP}(2)-\text{conv3D}(n,128)-\text{MP}(2)-\text{conv3D}(n,256)-\text{MP}(2)-\text{conv3D}(n,256)-\text{MP}(2)-\text{conv3D}(n,256)-\text{MP}(2)-\text{FC}(2048)-\text{FC}(2048)-\text{FC}(101)$. Each 3D convolutional and fully connected layer is followed by a ReLU activation function. All the convolution layers are applied with appropriate padding and stride 1 such that there is no change in size of the tensor from the input to the output of these layers. Following \cite{tran2015learning}, the input to the network are videos of dimension $3\times16\times112\times112$.

\begin{table}[b]   
	\centering
	\scriptsize
	\begin{tabular}[width=.9\textwidth]{l|c |c |c | c}
		\hline
		\hline
		\textbf{Network} & \textbf{Parameters}	& \textbf{Model Size} & \textbf{FLOP} & \textbf{Acc.}	\\
		& \textbf{(Millions)}	& \textbf{(Mb)} & \textbf{(Millions)} & \textbf{(\%)}	\\
		\hline
		2D-ResNet 18 \cite{he2016deep,tran2017convnet}			& $\approx$ 11.2	& -	& - & 42.2	\\
		2D-ResNet 34 \cite{he2016deep,tran2017convnet}			& $\approx$ 21.5	& -	& - &42.2	\\
		3D-ResNet 18  \cite{tran2017convnet}			& $\approx$ 33.2	& 254 	& - & 45.6	 \\
		3D-ResNet 34  \cite{tran2017convnet}			& $\approx$ 63.5	& 485	& - & 45.9	 \\
		3D-ResNet 101  \cite{diba2018spatio}		&  $\approx$ 86.06	& 657	& - & 46.7	 \\
		3D STC-ResNet 101 \cite{diba2018spatio}	&  -    & - & - & 47.9     \\
		\hline	
		$\text{mC3D}_3$ \cite{tran2015learning} (baseline)	&  $\approx$ 18	  &    139.6 & 34.88 & 44\\
		$\text{LP-mC3D}_{3}$ (ours)		      &  $\approx$ 13  & 106.2	& 26.072 &	\textbf{53.58} \\
		
		\hline	
		$\text{mC3D}_5$ \cite{tran2015learning} (baseline) 	&  $\approx$ 34.32 &  274.6	& 68.64 & 42.5	\\
		$\text{LP-mC3D}_{5}$  (ours) 		    &  $\approx$ 13   & 106.2		& 26.077	& 51.44  \\
		
		\hline	
		$\text{mC3D}_7$ \cite{tran2015learning} (baseline)	&  $\approx$ 71.88 & 575    & 143.72 &42.3	\\
		$\text{LP-mC3D}_{7}$  (ours) 		&   $\approx$ 13   & 106.2		& 26.08 & 50.54	\\
		
		\hline	
		$\text{mC3D}_9$  (baseline)	& $\approx$ 138.34	&    1100	& 276.68 &	36.17 \\
		$\text{LP-mC3D}_{9}$  (ours) 		&   $\approx$ 13   & 106.2		& 26.083 & 48.99 \\
		
		\hline
	\end{tabular}
	\caption{\textbf{Performance results on the UCF-101 split-1 action recognition dataset.} Comparison of the ReLPV block based 3D CNNs with their corresponding baselines and other state-of-the-art networks. All the networks are trained from scratch. }
	\label{tab:results_baseline_1}
\end{table}

The equivalent local phase version of the above network, denoted as $\text{LP-mC3D}_n$,  
is prepared by replacing the standard 3D convolutional layers with the ReLPV blocks as done in Section~\ref{sect:baselines}. Here, $n$ denotes  the size of the local 3D neighborhood in which STFT is computed. 

\noindent\textbf{Training.} Following \cite{tran2015learning}, we use SGD as optimizer with Nesterov momentum with value 0.9 and categorical crossentropy as loss. We train the networks for 16 epochs starting with a learning rate of 0.003 and decreasing it by a factor of 10 after every 4 epochs. Note that all the networks are trained from scratch. No data augmentation such as frame translation, rotation, or scaling is used. We re-trained all the baseline networks (for $n=3,5,7$). The results were found to be consistent with Fig.2 in \cite{tran2015learning}.

\noindent\textbf{Results.} Early works such as \cite{tran2015learning, karpathy2014large} showed that training relatively shallow 3D CNNs from scratch on the UCF-101 split-1 dataset achieve performance between $41-44\%$. Recent works such as \cite{tran2017convnet,diba2018spatio} use deep 3D Residual ConvNet architectures to achieve better results. Table~\ref{tab:results_baseline_1} reports our results on the UCF-101 split-1 dataset. We improve the state-of-the-art by 5.68\% while using just five ReLPV blocks. Our network uses 13 million parameters compared to the 3D STC-ResNet 101 network \cite{diba2018spatio}, which is built on the top of 3D ResNet 101 network and uses more than 86 million parameters. Furthermore, all the local phase versions with different local phase volumes significantly outperform the corresponding baseline networks.

\section{Discussion and Analysis}\label{sec:discussion}
In this section, we present detailed ablation and performance studies of the ReLPV block. Furthermore, we discuss some statistical advantages afforded by the ReLPV block over the standard 3D convolutional layer.

\subsection{Space-time Complexity of the ReLPV block}
\noindent\textbf{Model size.} Table~\ref{tab:results_baseline_1} shows that the ReLPV block based 3D CNNs use less parameters and  occupy less disk space when compared to the corresponding baselines. Furthermore, with an increase in the local phase volume (while keeping other hyperparameters constant) from $3$ to $9$, there is no change in the number of trainable parameters or model size in the ReLPV block based networks. In contrast, there is a significant rise in the number of parameters and model size in baseline networks with an increase in filter size. We believe this feature of the ReLPV block can be of huge benefit for 3D CNNs in resource constraint environment. 

\noindent\textbf{Computational cost.} We discussed in Section~\ref{sec:method} that due to the separability of the basis functions, STFT can be computed efficiently by using simple 1D convolutions for each dimension. This technique of computing 3D STFT using separable convolutions saves huge computational costs and has been of recent interest in 3D CNNs as discussed in Section~\ref{sec:related_work}. Table~\ref{tab:results_baseline_1} reports the computation cost in terms of the number of Floating Point Operations (FLOP) of the models. The FLOP values of the ReLPV block based 3D CNN are less when compared to the corresponding baselines. Furthermore, they vary very little with an increase in the local phase volume. However, for the baseline networks, the FLOP values increase by almost 8 times with an increase of filter size from 3 to 9.

\subsection{Statistical advantages of the ReLPV block}\label{sec:overfit}
\begin{figure}[t]
	\begin{center}
		\includegraphics[width=0.9\columnwidth, height=0.6\columnwidth]{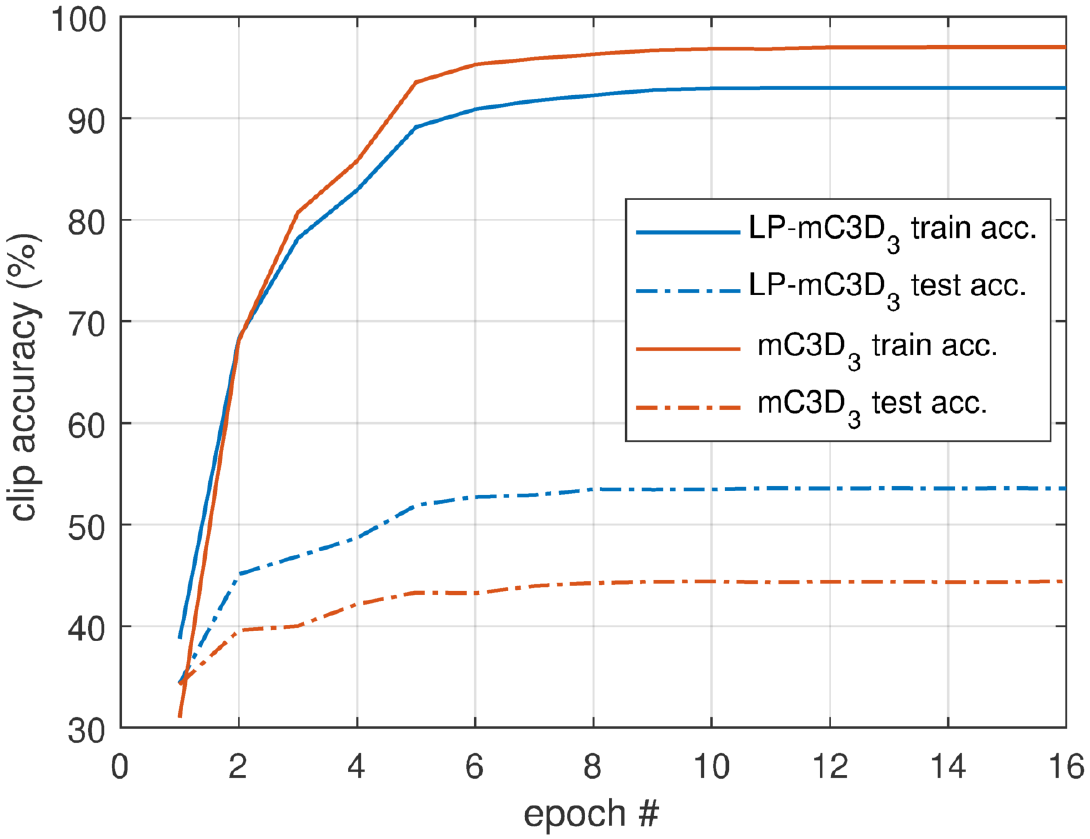} 
	\end{center}
\vspace{-1.5em}
	\caption{\textbf{Overfitting results on the UCF-101 split-1  dataset.} The $\text{LP-mC3D}_{3}$ network overfits less and generalizes significantly better compared to the baseline $\text{mC3D}_{3}$ network.}
	\label{fig:overfitting}
\end{figure}

As discussed earlier, one of the major challenges in training deep 3D CNNs is to avoid overfitting \cite{tran2015learning,tran2017convnet,hara2018can}. A recent study by Hara \emph{et al.} in \cite{hara2018can} shows that even a relatively shallow 3D CNN such as 3D ResNet-18 tends to overfit significantly on action recognition datasets such as UCF-101 \cite{soomro2012ucf101} and HMDB-51 \cite{kuehne2011hmdb}. This is partly due to the large number of trainable parameters in 3D CNNs in comparison to their 2D counterparts and partly due to the unavailability of large scale 3D datasets \cite{tran2017convnet,hara2018can}. These pose a major bottleneck in training deep 3D CNNs. In order to curb overfitting, various training methods such as data augmentation, training shallow networks,  and novel regularizers such as Dropout \cite{srivastava2014dropout}, DropConnect \cite{wan2013regularization}, and Maxout \cite{goodfellow2013maxout} have been introduced. While regularizers such as \cite{wan2013regularization, srivastava2014dropout, goodfellow2013maxout} have been proposed to regularize the fully connected layers of the network, recent works such as \cite{clevert2015fast,ioffe2015batch,srivastava2014dropout} show that regularizing the convolutional layers of the network is equally important. Our ReLPV block when used in the place of the standard 3D convolutional layer in deep 3D CNNs, naturally regularizes the network due to its use of significantly less trainable parameters and due to the decorrelation property of STFT (see Section~\ref{sec:method}). Fig.~\ref{fig:overfitting} reports our result on the overfitting experiment. The $\text{LP-mC3D}_{3}$, network clearly overfits less and generalizes significantly better when compared to the baseline $\text{mC3D}_{3}$ network.

\begin{figure}[t]
	\begin{center}
		\includegraphics[width=0.9\columnwidth, height=0.6\columnwidth]{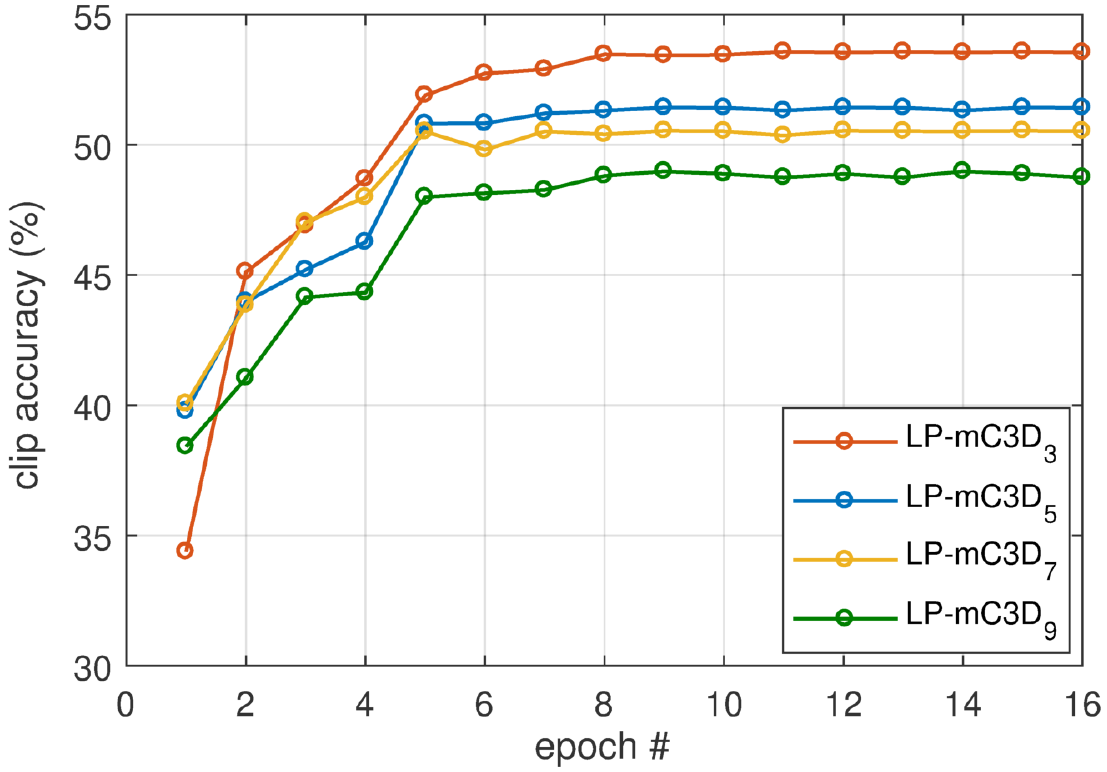} 
	\end{center}
	\vspace{-1.5em}
	\caption{\textbf{ReLPV block STFT volume search.} $\text{LP-mC3D}_3$ network with STFT volume of $3\times3\times3$ performs the best.}
	\label{fig:filter_size}
\end{figure}

%
%
%

\subsection{Exploring the Local Phase Volume of the ReLPV block}\label{sec:volume_search}
As described earlier, the ReLPV block takes two hyperparameters as input, one of which is the size of the local volume in which the STFT is computed (and the local phase is extracted) for each position of the input feature map. In this section, we explore this hyperparameter.  We experiment with different sizes of local volumes, in particular from $3\times3\times3$ to $9\times9\times9$. We found that the performance of the ReLPV block decreases with an increase in the STFT volume. Fig.~\ref{fig:filter_size} presents the clip accuracy of the $\text{LP-mC3D}_{n}$ network on the UCF-101 test split-1 dataset for various STFT volumes ranging from $3\times3\times3$ to $9\times9\times9$ over 16 epochs. The $\text{LP-mC3D}_{3}$ network with STFT volume of $3\times3\times3$ performs the best while the $\text{LP-mC3D}_{9}$ network performs the worst. Note that, an analogous study was carried out in \cite{tran2015learning} for the standard 3D convolutional layer where it was found that the 3D CNNs with $3\times3\times3$ convolutional kernels in all the layers perform the best.


\subsection{Exploring the number of feature maps output by the ReLPV block}
In this section, we explore another hyperparameter, the number of feature maps output by the ReLPV block. In simple words, we explore the effect of varying the number of $1\times1\times1$ filters in the \textit{Layer 4} of the ReLPV block (see Section~\ref{sec:method}). For this, we use a modified version of the $\text{LP-mC3D}_n$ network and experiment with different pairs of ReLPV block hyperparameters $(n,f)$. Let  $\text{LP-mC3D}_{n,f}$ be our experimental network with the following architecture: $\text{Input layer}- \text{ReLPV}(n,f)-\text{MP}(2)-\text{ReLPV}(n,f)-\text{MP}(2)-\text{ReLPV}(n,f)-\text{MP}(2)-\text{ReLPV}(n,f)-\text{MP}(2)-\text{ReLPV}(n,f)-\text{conv3D}(1\times1\times1,256)-\text{MP}(2)-\text{FC}(2048)-\text{FC}(2048)-\text{FC}(101).$ The layer $\text{Conv3D}(1\times1\times1,256)$ is used after the last ReLPV block so that the number of parameters in the fully-connected layers does not vary across different networks. Table~\ref{tab:feature_maps} presents our results of the experiment on the UCF-101 split-1 test set. We observe that, for a fixed value of the local STFT volume (the hyperparameter $n$), performance improves with an increase in the number of $1\times1\times1$ filters (the hyperparameter $f$). Another important observation is that the model size and the number of trainable parameters vary by a very small amount with an increase in the value of the hyperparameter $f$.

\subsection{ReLPV Block based Hybrid 3D CNN Models}
In this section, we explore the performance effects of using ReLPV blocks and the standard 3D convolutional layers in a single 3D CNN network. We call such networks as hybrid 3D CNNs. We experiment with two types of variations. In the first variation, we replace the top few layers (following the input layer) of a traditional 3D CNN network (baseline $\text{mC3D}_{3}$) with the ReLPV blocks such that the feature maps learned by the ReLPV blocks are input to the later standard 3D convolutional layers. In the second variation, the bottom layers are replaced with the ReLPV blocks such that the feature maps learned by the standard 3D convolutional layers are input to the later ReLPV blocks. We use the notation $\text{mC3D}_{3}(B_{l}/T_{l})$ to denote that $l$ bottom/top successive 3D conv layers of $\text{mC3D}_{3}$ are replaced with the ReLPV block. Table~\ref{tab:results_baseline} reports our results of the experiments. We observe that replacing standard 3D convolutional layers with the ReLPV blocks at the top of a traditional 3D CNN network improves its performance while the opposite happens when ReLPV blocks are added in the bottom layers. However, the hybrid 3D CNNs do not outperform the $\text{LP-mC3D}_{3}$ network where all layers are replaced with the ReLPV block (Table~\ref{tab:results_baseline_1}).  

\begin{table}[t]   
	\centering
	\scriptsize
	\begin{tabular}[width=.9\textwidth]{l|c | c |c}
		\hline
		\hline
		\textbf{Network} & \textbf{Parameters (Milions)} & \textbf{Model Size (Mb)}	 & \textbf{Acc.}	\\
		\hline
		$\text{LP-mC3D}_{3,64}$ 	     &  $\approx$ 12.84  & 104.2	&	50.96 \\
		$\text{LP-mC3D}_{3,128}$	     &  $\approx$ 12.93  & 104.9	& 51.84 \\
		$\text{LP-mC3D}_{3,256}$      &  $\approx$ 13.20  & 107.1	&	53.50 \\
		\hline	
		$\text{LP-mC3D}_{5,64}$   &  $\approx$ 12.84   	& 104.2	&	50.29		 \\
		$\text{LP-mC3D}_{5,128}$   &  $\approx$ 12.93  	& 104.9	&	51.10\\
		$\text{LP-mC3D}_{5,256}$   &  $\approx$ 13.20   & 107.1		&	53.22	\\
		\hline	
		$\text{LP-mC3D}_{7,64}$  	&   $\approx$ 12.84  & 	104.2	& 47.66	\\
		$\text{LP-mC3D}_{7,128}$  	&   $\approx$ 12.93  &	104.9	& 50.10 	\\
		$\text{LP-mC3D}_{7,256}$  	&   $\approx$ 13.20  &	107.1	& 51.14	\\
		\hline
	\end{tabular}
	
	\caption{\textbf{Exploring the number of feature maps output by the ReLPV block}. Performance improves with increase in value of $f$.}
	\label{tab:feature_maps}
\end{table}

\begin{table}[t]   
	\centering
	\scriptsize
	\begin{tabular}[width=.9\textwidth]{l|c |c |c}
		\hline
		\hline
		\textbf{Network} & \textbf{Parameters (Millions)} & \textbf{Model Size (Mb)}	 & \textbf{Acc.}	\\
		\hline
		$\text{mC3D}_{3}(T_1)$ 	     &  $\approx$ 17.44  & 139.9	&	51.51 \\
		$\text{mC3D}_{3}(T_2)$	     &  $\approx$ 16.13  & 138.5	& 47.67 \\
		$\text{mC3D}_{3}(T_3)$      &  $\approx$ 13.20  & 132.1	&	43.95 \\
		\hline	
		$\text{mC3D}_{3}(B_1)$ 	     &  $\approx$ 15.82  & 126.9	&	35.1 \\
		$\text{mC3D}_{3}(B_2)$	     &  $\approx$ 14.13  & 113.7	& 36.47 \\
		$\text{mC3D}_{3}(B_3)$      &  $\approx$ 13.30  & 107.3	&	40.84 \\
		\hline	
	\end{tabular}
	\caption{\textbf{Results on hybrid 3D CNN architectures}. Performance results on the UCF-101 split-1 test set.}
	\label{tab:results_baseline}
\end{table}

\section{Conclusion}\label{sec:conclusion}
In this work, we have proposed ReLPV block, an efficient alternative to the standard 3D convolutional layer, in order to reduce the high space-time and model complexity of the traditional 3D CNNs. The ReLPV block when used in place of the standard 3D convolutional layer in traditional 3D CNNs, significantly improves the performance of the baseline architectures. Furthermore, they produces consistently better results across different 3D data representations. Our proposed ReLPV block based 3D CNN architectures achieve state-of-the-art results on the ModelNet and UCF-101 split-1 action recognition datasets. We plan to apply ReLPV block in 3D CNN architectures for other 3D data representations and tasks such as 3D MRI segmentation.\\

\noindent\textbf{Acknowledgments.} The authors gratefully acknowledge the travel grant support from Google Research India. Sudhakar Kumawat was supported by TCS Research Fellowship. Shanmuganathan Raman was supported by SERB Core Research Grant and Imprint 2 Grant. We thank Manisha Verma for early contribution and technical discussion.

\appendix
\section{Detailed Mathematical Formulation of \textit{Layer 2} of the ReLPV Block}\label{sect:supp}

In this section, we elaborate on each step of the \textit{Layer 2} of ReLPV block which is at the core of our ReLPV block. 

Let $f(\textbf{x})$ be the single channel feature map of size $1\times d\times h\times w$ that is output by \textit{Layer 1} of the ReLPV block. Here, $h$, $w$, and $d$ denotes the height, width, and depth of the feature map, respectively.  For simplicity, we will drop the channel dimension and rewrite the size of $f(\textbf{x})$ as $ d\times h\times w$. Here, $\textbf{x}\in\mathbb{Z}^3$ are the 3D coordinates of the elements in $f(\textbf{x})$.

Every $\textbf{x}$ in $f(\textbf{x})$  has a $n\times n\times n$ 3D neighborhood denoted by $\mathcal{N}_{\textbf{x}}$ which is defined in Equation~\ref{eq:1}. We provide detailed experimental analysis in the manuscript on the effect of varying $n$ on the performance of the ReLPV block in 3D CNNs meant for video classification task.

\begin{equation}\label{eq:1}
\mathcal{N}_{\textbf{x}}=\{\textbf{y}\in\mathbb{Z}^3 \,; \parallel(\textbf{x}-\textbf{y})\parallel_{\infty} \leq r \, ; n=2r+1; r\in\mathbb{Z}_{+}\}
\end{equation}

For all positions $\textbf{x}=\{\textbf{x}_1,\textbf{x}_2,\ldots,\textbf{x}_{d\cdot h\cdot w}\}$ of the feature map $f(\textbf{x})$, we use local  3D neighborhoods, $ f(\textbf{x}-\textbf{y}) ,\forall \textbf{y}\in \mathcal{N}_{\textbf{x}}$ to derive the local frequency domain representation using Short Term Fourier Transform (STFT) as defined in Equation~\ref{eq:2}.

\begin{equation}\label{eq:2}
F(\textbf{v},\textbf{x})= \sum_{\textbf{y}_i\in \mathcal{N}_\textbf{x}} f(\textbf{x}-\textbf{y}_i)\exp^{-j2\pi \textbf{v}^T \textbf{y}_i}
\end{equation}

Here $i = 1,\ldots,n^3$, $\textbf{v}\in\mathbb{R}^3$ is a 3D frequency variable, and $j=\sqrt{-1}$. Using vector notation \cite{jain1989fundamentals}, we can rewrite Equation~\ref{eq:2} as shown in Equation~\ref{eq:3}.  

\begin{equation}\label{eq:3}
F(\textbf{v},\textbf{x})=\textbf{w}^T_\textbf{v}\textbf{f}_\textbf{x}
\end{equation}

Here,  $\textbf{w}_\textbf{v}$ is a complex valued basis function (at frequency variable $\textbf{v}$ ) of a linear transformation, and is defined as shown in Equation~\ref{eq:4}.

\begin{equation}\label{eq:4}
\textbf{w}^T_\textbf{v}=[\exp^{-j2\pi\textbf{v}^T\textbf{y}_1}, \exp^{-j2\pi\textbf{v}^T\textbf{y}_2},\ldots, \exp^{-j2\pi\textbf{v}^T\textbf{y}_{n^3}}] ,	
\end{equation}

and $\textbf{f}_\textbf{x}$ is a vector containing all the elements from the neighborhood $\mathcal{N}_\textbf{x}$, and is defined as shown in Equation~\ref{eq:5}.

\begin{equation}\label{eq:5}
\textbf{f}_\textbf{x}=[f(\textbf{x}-\textbf{y}_1), f(\textbf{x}-\textbf{y}_2),\ldots, f(\textbf{x}-\textbf{y}_{n^3})]^T
\end{equation}

In our work, we consider 13 lowest non-zero frequency variables $\textbf{v}_1, \textbf{v}_2,\ldots,\textbf{v}_{13}$.  Low frequency variables are used because they usually contain most of the information, and therefore they have better signal-to-noise ratio than the high frequency components \cite{heikkila2009methods} (see Section~\ref{sec:decorr}). The values of these frequency variables are already discussed in the main paper. Thus, from Equation~\ref{eq:3}, the local frequency domain representation for the above frequency variables is defined as shown in Equation~\ref{eq:6}.

\begin{equation}\label{eq:6}
\textbf{F}_{\textbf{x}} = [F(\textbf{v}_1,\textbf{x}), F(\textbf{v}_2,\textbf{x}),\ldots,F(\textbf{v}_{13},\textbf{x})]^T 
\end{equation}

At each position $\textbf{x}$, after separating the real and imaginary parts of each component, we get a vector as shown in Equation.~\ref{eq:7}.

\begin{multline}\label{eq:7}
\textbf{F}_{\textbf{x}} = [\Re\{F(\textbf{v}_1,\textbf{x})\}, \Im\{F(\textbf{v}_1,\textbf{x})\},\Re\{F(\textbf{v}_2,\textbf{x})\},\\ \Im\{F(\textbf{v}_2,\textbf{x})\},
\ldots,\Re\{F(\textbf{v}_{13},\textbf{x})\}, \Im\{F(\textbf{v}_{13},\textbf{x})\}]^T
\end{multline}

Here, $\Re\{\cdot\}$ and $\Im\{\cdot\}$  return the real and  imaginary parts of a complex number, respectively. The corresponding $26\times n^3$ transformation matrix can be written as shown in Equation~\ref{eq:8}.

\begin{equation}\label{eq:8}
\textbf{W} = [\Re\{\textbf{w}_{\textbf{v}_1}\}, \Im\{\textbf{w}_{\textbf{v}_1}\}, \ldots, \Re\{\textbf{w}_{\textbf{v}_{13}}\}, \Im\{\textbf{w}_{\textbf{v}_{13}}\}]^T
\end{equation}

Hence, from Equation~\ref{eq:3} and~\ref{eq:8}, the vector form of STFT for all the 13 frequency points $\textbf{v}_1, \textbf{v}_2,\ldots,\textbf{v}_{13}$ can be written as shown in Equation~\ref{eq:9}.

\begin{equation}\label{eq:9}
\textbf{F}_\textbf{x}=\textbf{Wf}_\textbf{x}
\end{equation}

Since, $\textbf{F}_\textbf{x}$ is computed for all positions $\textbf{x}$ of the input $f(\textbf{x})$, it results in an output feature map with size $26\times d\times  h\times w$. This feature map is then passed as input to the \textit{Layer 3} of the ReLPV block.\\


\section{Decorrelation Property of STFT and Reason for Selecting Low Frequency Variables}\label{sec:decorr}

As mentioned in the manuscript, some important properties of Short Term Fourier Transform (STFT) is its ability to decorrelate the input signal and to compact the energy (information) contained in a signal. These properties are inherent to STFT since it belongs to the family of orthogonal transforms such K-L transform, Walsh-Hadamard transform (WHT), and Discrete Cosine Transform (DCT) \cite{wang2012introduction}. All the above orthogonal transforms have the following properties in common.

\begin{itemize}
	\item Orthogonal transforms have the tendency of decorrelating the input signals \cite{wang2012introduction}. For example, consider a signal containing temperature as a function of time. Now, given the value of a current sample of the signal, the value of its next sample can be predicted with reasonable confidence to be close to the current one, i.e., two consecutive time samples are highly correlated. On the other hand, after an orthogonal transform, such as Fourier transform, knowing the magnitude of a certain frequency component, one has little idea in terms of the magnitude (or the energy) of the next frequency component, i.e., the two components are much less correlated than the time samples before the transform. The same property holds true for signals in multiple dimensions such as images and videos \cite{hinman1984short}. In images and videos, decorrelation is achieved due to STFT's insensitivity  to the correlation coefficient of images and videos \cite{hinman1984short}.
	
	\item Orthogonal transforms tend to compact the energy (information) contained in the signal into a small number of signal components \cite{wang2012introduction}. For example, after Fourier transform, most of the energy (information) will be concentrated in  a relatively small number of low frequency components. Most of the high frequency components carry little energy. Moreover, low frequency components have better signal-to-noise ratio than the high frequency
	components. It is for this reason that we chose low frequency variables while computing STFT.  
\end{itemize}

\end{document}